\pdfoutput=1
\documentclass[10pt]{article}
\usepackage[preprint]{tmlr}

\usepackage{hyperref}
\usepackage{url}
\usepackage{booktabs}
\usepackage{array}
\usepackage{graphicx}
\usepackage{xcolor}
\usepackage{amsmath}
\usepackage{amsfonts}

\title{Revisiting Ground-Truth Synthesis from High-Speed Video:\\
Exact Validity Conditions and an Audited\\ Consumer Capture Corpus}

\author{\name Abdullah Al Shafi, Sumaiya Rahim Suma \\
      \addr Department of Computer Science and Engineering\\
      Khulna University of Engineering \& Technology, Khulna, Bangladesh\\
      \{abdullah.shafi99, sumaiya.rahim234\}@gmail.com}

\def\month{MM}
\def\year{YYYY}
\def\openreview{\url{https://openreview.net/forum?id=XXXX}}

\renewcommand{\headrulewidth}{0pt}

\begin{document}
\maketitle

\begin{abstract}
Motion-deblurring datasets are commonly synthesised by averaging $N$ consecutive
high-frame-rate frames and labelling the result with the central frame. We show that this label is unbiased for every capture timing only when the window is
odd and the frames' sample durations are equal. An even window shifts every label by a fixed
fraction of the blur length, even under perfect timing. Unequal durations are
subtler: their mean misalignment is zero, so dataset statistics cannot reveal them,
yet when the offset cannot be read from the blur they convolve the supervision
rather than adding noise to it. Auditing 51 smartphone clips recorded at a nominal 240\,fps, we find 19 captured
near 176\,fps, a behaviour recorded only in the container's timing tables. In a controlled test the parity choice costs a fitted linear deblurring filter far more than
these timing irregularities do, and interpolation labels, unlike blur labels, can
be repaired with the true frame times. We provide a tool that checks both conditions without decoding, and will release
the clips and their timing tables.
\end{abstract}

\section{Introduction}
\label{sec:intro}

Most real-world motion-deblurring data is manufactured rather than captured. The
recipe of \citet{nah2017gopro} and \citet{su2017dvd} records at a high frame rate,
averages $N$ consecutive frames to approximate a long exposure, and designates the
central frame as the sharp target. The same recipe still underlies new datasets
\citep{lee2024gsblur,mahmud2025slomodeblur}. Both images are synthetic, and their
pairing rests on an assumption that is rarely stated: that the central frame is
where the exposure is centred.

\begin{figure}[!htb]
\centering
\includegraphics[width=0.80\linewidth]{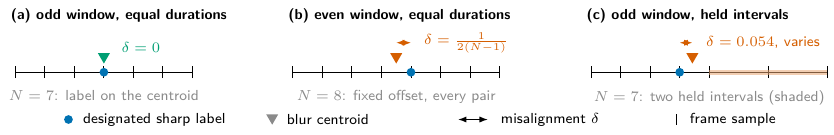}
\caption{The sharp label against the centroid of the blur it is paired with.
(a)~Odd window, equal durations: aligned. (b)~Even window: a fixed offset in every
pair. (c)~Held intervals: an offset that varies and averages to zero.}
\label{fig:mechanism}
\end{figure}

We give the exact condition under which it is. Writing $\delta$ for the
displacement between the sharp label and the intensity centroid of the
synthesised blur, normalised by the blur length, we show that $\delta$ vanishes
exactly when a parity term and a timing term cancel (Eq.~\ref{eq:weights}). With
equal sample durations the timing term is zero and
\begin{equation*}
\delta = 0 \quad\Longleftrightarrow\quad N \text{ is odd.}
\end{equation*}
An odd window of equal durations is therefore the configuration that guarantees
an unbiased label, and both properties can be checked before any synthesis. Each
is violated in practice, differently (Figure~\ref{fig:mechanism}).

The first is violated by choice. An even window has no central frame, so every
pair is offset by $1/(2(N-1))$ of the blur length however perfect the capture,
$7.1\%$ at $N=8$, in the same direction every time
(Figure~\ref{fig:mechanism}b). Published pipelines use both parities
(Section~\ref{sec:related}).

The second is violated by hardware. On the smartphone we audit, 19 of 51 clips
recorded at a nominal 240\,fps were captured near 176\,fps, holding about a third of
their intervals for two slots, and only the container's sample-duration table
reveals it (Section~\ref{sec:corpus}). The resulting misalignment has zero mean
(Figure~\ref{fig:mechanism}c), harmless for noise on a target's \emph{value}
\citep{lehtinen2018noise2noise} but not for a perturbation of its position
\emph{in time}: the expected label is the sharp frame convolved with the jitter
density, so the supervision is not noisy, it is blurred.

This paper makes four contributions. It states the exact validity condition and the misalignment each violation causes
(Section~\ref{sec:model}). It shows that zero-mean timing error convolves the
supervision, with a controlled test separating the costs of the two violations
(Section~\ref{sec:tests}). It audits iPhone-HighFPS, the 51-clip corpus in which the hold behaviour was found
and which we will release (Section~\ref{sec:corpus}). And it traces the
consequences for deblurring and interpolation labels, with a tool that checks both
conditions before a dataset is built (Section~\ref{sec:consequences}).

\paragraph{Scope of the claims.}
The validity condition is exact for any capture under the locally linear motion
model; the audit covers one
device across two recording sessions (Section~\ref{sec:limitations}). The odd-$N$ identity is the antisymmetry argument behind the midpoint quadrature rule
\citep{davis1984numerical}, and we know of no prior statement of the centroid
condition in this setting; what we claim is its statement, quantification and
consequences here. We do
not claim the capture behaviour is a defect, that we have identified its trigger,
or that any published benchmark's numbers are invalidated.

\section{Related work}
\label{sec:related}

\paragraph{The averaging convention.}
\citet{nah2017gopro} and \citet{su2017dvd} established it; \citet{su2017dvd}
interpolate intermediate frames to avoid ghosting from the gap between exposures,
and \citet{nah2019reds} upsample the frame rate before averaging.
\citet{rim2022realistic} add saturation, noise and camera processing to the
synthesis, and real-captured datasets avoid it altogether
\citep{rim2020realblur,zhong2020bsd}. The synthetic pipelines all presuppose uniformly spaced source frames, and none checks
that they are.

\paragraph{Window parity in practice.}
Published pipelines fall on both sides of the condition, rarely saying why.
\citet{jin2018learning} use seven frames rather than eight to avoid ambiguity in
the middle frame, the right window for a different reason; GoPro averages 7 to 13
frames around the mid-frame \citep{nah2017gopro}, and a rendered dataset takes the
middle of 121 sub-frames \citep{lee2024gsblur}. A recent smartphone benchmark
averages $N = 30$ frames with frame $\lfloor N/2 \rfloor$ as reference
\citep{mahmud2025slomodeblur}, by Eq.~\ref{eq:even} an offset of $1/58 = 0.017$ of
the blur length, about $0.7$ pixels on a 40-pixel streak, under perfect timing. The
same work describes the phone's high-speed pipeline as offering superior
frame-timing stability, which Section~\ref{sec:populations} qualifies for the device
we audit. An infrared study takes frame $N/2+1$ for even $N$ up to 13
\citep{varghese2021ir}, offsets of $0.167$, $0.100$ and $0.071$ at $N = 4$, 6 and 8.
We raise these cases to show the condition binds in practice, not to question any
work's results. 

\paragraph{Temporal labels and pipeline audits.}
Interpolation benchmarks assign each ground-truth frame a normalised time
\citep{xue2019vimeo,choi2020cain}. \citet{zhong2024clearer} replace time with a
distance-based index to resolve velocity ambiguity and \citet{xu2019qvi} model
acceleration, both assuming the timestamps are right; our concern is whether they
are. Dedicated high-speed rigs \citep{sim2021xvfi,ciubotariu2025miore} lie outside
the consumer regime, \citet{ji2023rethinking} note that discontinuous averaging
produces unnatural motion artifacts, and \citet{shang2023selfdrsc} model per-row
scanning time, which is intra-frame. Our work joins audits showing that low-level
pipeline choices silently bias benchmarks
\citep{parmar2022aliased,northcutt2021pervasive}. Closest in domain,
\citet{kiefhaber2024benchmarking} question the \emph{content} of interpolation
ground truth where we question its \emph{construction}, and the \citet{acevfi2025}
survey has no timing-integrity axis. The container structures we read are standard
in multimedia forensics \citep{gloe2014forensic,iuliani2019video,yang2020container}.
The vendor documents one automatic frame-rate reduction, Auto FPS, which lowers 30
and 60\,fps video to 24\,fps in low light \citep{apple2026autofps}, and none for
high-frame-rate capture.

\section{When is the central-frame label unbiased?}
\label{sec:model}

A frame-averaged label is unbiased exactly when the averaged instants are
symmetric about it; four propositions make this exact and quantify what each violation costs. Consider a window of $N$ frames with presentation times $T_0,\dots,T_{N-1}$,
averaged with equal weight $1/N$, with the frame at $m = \lfloor N/2 \rfloor$
designated the sharp reference, and span $S = T_{N-1} - T_0$. Under locally
linear motion the blur centroid lies at the mean sample time $\bar{T}$. Define the
signed misalignment $\tilde{\delta} = (T_m - \bar{T})/S$ and its magnitude
$\delta = |\tilde{\delta}|$.

With $\Delta$ the nominal frame interval, write the durations as
$d_j = \Delta(1 + b_j)$, where $b_j$ counts the extra nominal slots interval $j$
occupies. Then
\begin{equation}
T_m - \bar{T} \;=\;
\underbrace{\Delta\left[m - \tfrac{N-1}{2}\right]}_{\text{parity term}}
\;+\;
\underbrace{\Delta \sum_{j=0}^{N-2} b_j\, w_j}_{\text{timing term}},
\qquad
w_j \;=\; \mathbf{1}[j < m] - \frac{N-1-j}{N},
\label{eq:weights}
\end{equation}
with $S = \Delta\,(N - 1 + \sum_j b_j)$. The parity term depends only on window
size and survives perfect timing; the timing term depends only on where held
intervals fall. Since $\sum_j w_j = m - \tfrac{N-1}{2}$, $\delta = 0$ exactly when
$\sum_j b_j w_j = \tfrac{N-1}{2} - m$. Normalising by the exposure span $N\Delta$ instead of $S$ would scale $\delta$ by
$(N - 1 + \sum_j b_j)/N$, which is $(N-1)/N$ under perfect timing.

\paragraph{P1: the odd-window condition.}
Set every $b_j = 0$ to recover uniform spacing, leaving only the parity term.
It vanishes exactly when $N$ is odd, so the reference sits at the centroid and
$\delta = 0$. For even $N$, $m = N/2$ and the parity term equals $\Delta/2$;
dividing by $S = \Delta(N-1)$ gives
\begin{equation}
\delta_{\text{even}} \;=\; \frac{1}{2(N-1)}
\label{eq:even}
\end{equation}
under perfect timing. The bias is \emph{systematic} in time: every pair is
misaligned the same way by the same fraction of its blur length, so it survives
dataset averaging. In the image, though, a streak from locally linear motion does
not reveal which way the motion ran \citep{jin2018learning,purohit2019bringing},
so under $L_2$ the optimum averages the label shifted both ways: a faint double
image rather than a consistent shift.

\paragraph{P2: zero mean does not mean harmless.}
Take $N$ odd, so only the timing term remains. If held intervals fall with
probability $p$ then $\mathbb{E}[T_m - \bar{T}] = \Delta p \sum_j w_j
= \Delta p\,[m - (N-1)/2] = 0$: the offset averages to zero over a dataset, which needs
only a constant marginal hold probability across the window, not independence. The
normalised $\tilde{\delta}$ also has zero mean when hold patterns are as likely
reversed as forwards, since $w_{N-2-j} = -w_j$ for odd $N$ and $S$ is unchanged. Under $L_2$ the optimum is unchanged when a target is replaced by
any distribution with the same conditional expectation
\citep{lehtinen2018noise2noise}, so zero-mean noise on the target's \emph{value}
would be benign. Here the label is the frame at a perturbed time, not the correct
frame with perturbed intensity, so with $I(t)$ the scene and $\varepsilon$ the
temporal displacement,
\begin{equation}
\mathbb{E}_{\varepsilon}\!\left[ I(\bar{T} + \varepsilon) \right]
\;=\; \left( I * \check{p}_{\varepsilon} \right)\!(\bar{T}),
\qquad \check{p}_{\varepsilon}(e) = p_{\varepsilon}(-e),
\label{eq:convolve}
\end{equation}
the clean frame convolved with the reflected jitter density: the $L_2$ optimum is a
blurred target.

\paragraph{P3: the scale in closed form.}
If holds fall independently,
$\operatorname{Var}[\sum_j b_j w_j] = p(1-p)\sum_j w_j^2$, and for odd $N$
\begin{equation}
\sum_{j} w_j^2 \;=\; \frac{m(m+1)(2m+1)}{3N^2}, \qquad m = \tfrac{N-1}{2},
\label{eq:energy}
\end{equation}
so a builder who measures $p$ can estimate the misalignment their synthesis will
incur, to order of magnitude, without any per-sample computation.

\paragraph{P4: a bound on the supervision.}
By Eq.~\ref{eq:convolve}, jitter of standard deviation $\sigma_\delta$ convolves the
label with a kernel of width $\sigma_\delta L$ for blur length $L$ in pixels. For any estimator that cannot read the offset from the blurred input, this limits
the sharpness available in the supervision regardless of sample size. A hold
leaves a gap in the streak that could make its offset readable; a linear filter
fitted under random direction is symmetric and cannot express such a shift, so the
test of Section~\ref{sec:tests} shows only that the hold offset is not
\emph{linearly} recoverable. The mechanism belongs to the errors-in-variables family
\citep{fuller1987measurement}, except that the conditional
mean is blurred rather than shrunk toward zero.

\paragraph{The validity condition.}
Unbiasedness requires the timing term to cancel the parity term, which under equal
durations means $N$ odd. Unequal durations break it for either parity except in
balanced cases, such as $N = 5$ with the first and last intervals held (times
$0, 2, 3, 4, 6$ in units of $\Delta$, mean $T_2 = 3$), which a builder cannot control at capture time, so the guarantee requires both
properties. The even-window offset keeps
its sign and survives averaging; the timing offset has zero mean and does not, so
the quantity to report is the spread rather than the mean.

\subsection{What the misalignment costs a fitted estimator}
\label{sec:tests}

The two violations should cost a learned estimator very different amounts. On identical inputs, an eight-frame window costs a fitted linear deblurring filter
$0.42$\,dB at 16 to 32 pixels of blur, about eight times the $0.05$\,dB that real hold
patterns cost a seven-frame window at the same blur lengths
(Table~\ref{tab:controlled}).

\begin{table}[!htb]
\caption{What each violation costs a fitted linear deblurring filter: PSNR loss of
the conventional label against the frame at the true centroid, mean and standard
deviation over five seeds, with the net shift the filter learns.}
\label{tab:controlled}
\begin{center}
\begin{tabular}{llll}
\toprule
Violation & Blur length & Cost (dB) & Learned shift \\
\midrule
Even window, random direction & 16--32 px & $0.42 \pm 0.05$ & none: offset becomes blur \\
Even window, random direction & 32--48 px & $0.73 \pm 0.07$ & none \\
Even window, direction fixed  & 16--32 px & $0.85 \pm 0.03$ & 1.08 of the 1.72-px offset \\
Real hold patterns            & 8--48 px  & $0.04$ to $0.05$ & none \\
\bottomrule
\end{tabular}
\end{center}
\end{table}

Synthetic scenes, crops of standard test images split into disjoint training and
test halves, translate horizontally at constant speed, the direction drawn at
random so the model cannot know it. We fit the $L_2$-optimal linear filter to the conventional label and to the frame at
the true centroid: a $3 \times 81$ ridge fit on 150,000 pixels from 500 windows, its
support wider than the longest blur it must invert. Both are scored on the same 200
held-out windows against the centroid frame, rendered directly even when it falls
between sampled frames.

\begin{figure}[!htb]
\begin{center}\includegraphics[width=0.80\linewidth]{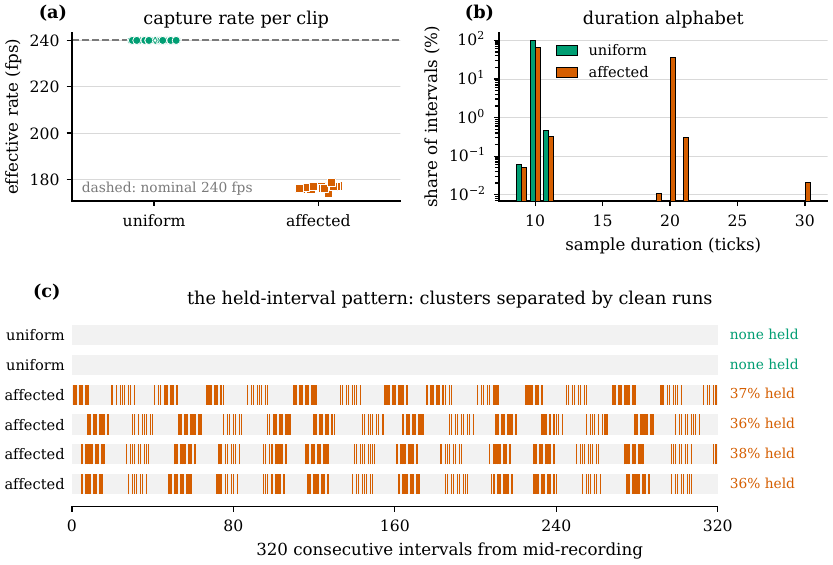}\end{center}
\caption{The corpus at a glance. (a)~Effective capture rate per clip; the dashed line is
the nominal 240\,fps, which uniform clips meet and affected clips miss at
174--179\,fps. (b)~Share of intervals at each sample duration (log axis): one
ten-tick slot for uniform clips; a second mode at twenty ticks, exactly twice
nominal, for affected ones. The $\pm1$ tick spread is quantisation dither.
(c)~Held intervals, one mark each, in 320 consecutive mid-recording intervals of four
affected and two uniform clips: clusters separated by clean runs, and none in
uniform clips.}
\label{fig:overview}
\end{figure}

Under random direction the filter learns no net shift, so the parity offset becomes
blur, exactly as P1 predicts. Fixing the direction makes the offset learnable, and the filter reproduces most of
it, moving its output off the centroid, short of the whole because the ridge penalty and
the $L_2$ objective trade shift against sharpness. The hold patterns of the four affected clips of Figure~\ref{fig:overview}c, transplanted onto the scenes, cost little, with a seed standard deviation of about $0.02$\,dB and a positive cost in
every seed, and the filter recovers none of the hold-induced offset.

These costs are measured against a correctly centred label. A dataset built with
an even window trains and evaluates on labels carrying the same offset, so, as with
the aggregation in P2, its own metrics report none of it. Whether real capture produces held intervals, instants the sensor never captured,
is a question the container's timing table can answer without decoding a frame.

\section{iPhone-HighFPS: an audited consumer capture corpus}
\label{sec:corpus}

Read directly from the container, the timing tables split a corpus of consumer
clips into two populations, and only one satisfies the validity condition.

\subsection{Corpus and timing extraction}

The corpus, which we will release as iPhone-HighFPS, is 51 camera-original QuickTime
files from one smartphone (iPhone~17~Pro, iOS~26.0), HEVC at $1920\times1080$ and
nominal 240\,fps: 60.4 minutes and 741{,}919 frames (23.7\,GB) of handheld outdoor
urban footage from two sessions. No file carries an encoder fingerprint indicating
re-encoding, vendor camera metadata is intact on all 51, and both populations carry
identical writing metadata, so the transfer route cannot explain their difference.
The same footage also underlies the iPhoneBlur deblurring benchmark
\citep{shafi2026iphoneblur}, distinct from \citet{mahmud2025slomodeblur}, which
reports capture rates below the nominal 240\,fps.

We read timing from the \texttt{stts} atom, which stores exact integer sample
durations, so nothing is decoded. The timescale is 2400 ticks per second (one tick
is $416.7\,\mu$s, $10\%$ of a frame interval), and presentation times add signed
\texttt{ctts} offsets from $-41$ to $+31$ ticks; decode order gives identical
results. The negative offsets are permitted by QuickTime \citep{apple2026qtff} but
excluded from a version-0 \texttt{ctts} by the ISO format \citep{iso2022bmff}, so an ISO-strict parser reads each negative offset as unsigned and shifts that
sample by about $2^{32}$ ticks. Edit lists,
present in all 51 clips, do not alter the analysed range, and frames must be
extracted without resampling to a constant rate, which silently duplicates and
drops them. Rounding to whole ticks makes even a constant rate dither, so we
separate rounding from genuine variation by the residual of cumulative timestamps from a linear fit. A clip is uniform when its residual is at most $0.75$ ticks and non-uniform above
$1.5$, and no clip falls between. An interval is \emph{held} when it exceeds $1.5$ times the clip's median, which separates one ten-tick slot from two. Two frames are near-duplicates when their mean absolute difference falls below $1.0$ at 8-bit
depth, a bimodal statistic with no pair within an order of magnitude of the threshold.

\subsection{Two capture populations}
\label{sec:populations}

\begin{figure}[!htb]
\begin{center}\includegraphics[width=0.80\linewidth]{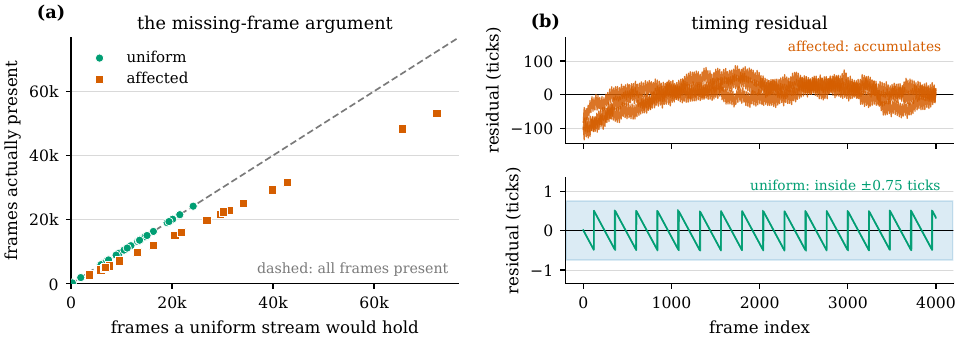}\end{center}
\caption{Why the timing is real. (a)~Frames present in each clip against the number a
uniform 240\,fps stream of the same duration would hold: affected clips fall well
below the identity line, so frames are missing. (b)~Residual of cumulative
timestamps from a linear fit, three clips per population on separate scales:
affected residuals wander far outside the $\pm0.75$ tick band (shaded); uniform
ones are the bounded sawtooth of dither.}
\label{fig:timing}
\end{figure}

\begin{table}[!htb]
\caption{The corpus separates under identical device, operating system and nominal
settings. Residuals are the maximum absolute deviation of cumulative timestamps
from a linear fit, in ticks; rounding alone stays near half a tick. Frame counts
sum to the corpus total of 741{,}919.}
\label{tab:populations}
\begin{center}
\begin{tabular}{lrr}
\toprule
 & \textbf{Uniform} & \textbf{Affected} \\
\midrule
Clips / duration / frames        & 32 / 26.8\,min / 385{,}263 & 19 / 33.7\,min / 356{,}656 \\
Sample durations (ticks)         & $\{9,10,11\}$ & $\{9,10,11,19,20,21,30\}$ \\
Share at one / two nominal slots & 100.00\% / --- & 64.16\% / 35.82\% \\
Effective rate & 239.90\,fps & 174--179\,fps \\
Linear-fit residual (ticks)      & 0.52--0.53 & 47.97--417.26 \\
\bottomrule
\end{tabular}
\end{center}
\end{table}

Table~\ref{tab:populations} summarises the separation, clean on every timing measure
and absent from every visual one, and Figure~\ref{fig:overview} shows it per clip.
Uniform durations take the values 9, 10 and 11 ticks, 10 accounting for $99.46\%$,
and the residual never exceeds $0.53$ ticks: a constant 239.90\,fps dithered onto
integer durations. In the affected population 20 ticks, exactly twice nominal, accounts for
$35.51\%$ of intervals; grouped by whole slots, $64.16\%$ occupy one, $35.82\%$ two
and $0.02\%$ three, which is why the model uses $b_j \in \{0,1\}$. Residuals reach
417 ticks (Figure~\ref{fig:timing}b). Durations fall exactly on their slot value at nearly the same rate in both
populations ($99.46\%$ against $99.31\%$), so the $\pm1$ tick dither is a property of
the clock and the holds are not.

\paragraph{The missing frames.}
An uneven table could also come from a container that padded a slower stream with
repeated frames, or that mislabelled the durations of a uniform one. An independent
clock rules out both. A second media track in the same file, timed on its own clock, matches the video
track's duration to within 2\,ms in every clip, about half a frame interval. Over
that duration the affected clips hold 356{,}656 frames where 240\,fps would give
484{,}752 (Figure~\ref{fig:timing}a). A representative affected clip holds 4{,}318 frames in 24.6 seconds where 240\,fps
would give 5{,}904, a deficit of 1{,}586 frames. Repeated frames and mislabelled durations each make further predictions,
and the data contradict every one (Table~\ref{tab:evidence}). The sensor therefore
delivered about 176 distinct frames per second, recorded by doubling durations on a
clock quantised to the nominal rate.

\subsection{What the populations are not explained by}
\label{sec:explanations}

Every affected clip holds between $34.2\%$ and $37.9\%$ of its intervals, mean
$35.8\%$, and every uniform clip exactly $0.0\%$: the populations correspond to
the two recording sessions, with no intermediate case. Ordinary explanations for an
uneven timing table each make a prediction, and Table~\ref{tab:evidence} sets every prediction against the measurement. None is supported by these comparisons. The state is decided 1.3 to 6.1 seconds into a recording (median 1.8) and then
held, matching the time auto-exposure and autofocus take to settle, which we note
as a hypothesis. The documented Auto FPS lowers 30 to 24\,fps, a factor of $0.80$ against the $0.73$ here, but only in low light, and the affected
clips are, if anything,
the brighter ones.

\begin{table}[!htb]
\caption{Evidence and ruled-out explanations. Each row is a hypothesis that would
produce an uneven timing table, what it predicts, and what the corpus shows.
Session-level comparisons are descriptive: with two sessions, clip-level tests
carry no session-level weight.}
\label{tab:evidence}
\begin{center}
\small
\setlength{\tabcolsep}{4pt}
\begin{tabular}{>{\raggedright\arraybackslash}p{0.16\linewidth}>{\raggedright\arraybackslash}p{0.19\linewidth}>{\raggedright\arraybackslash}p{0.43\linewidth}>{\raggedright\arraybackslash}p{0.13\linewidth}}
\toprule
Hypothesis & It predicts & The corpus shows & Verdict \\
\midrule
\addlinespace[2pt]
Padded stream & repeated, near-empty frames & $0.00\%$ near-duplicates; no frame under $2\%$ of median size; post-hold frames no smaller (ratio $1.10$) & excluded \\
\addlinespace[2pt]
Mislabelled durations & clock disagreement; motion ratio 1 & clock agrees to 2\,ms; motion across holds $1.72\times$ normal (IQR $1.59$--$1.94$, nine clips with flow above $0.05$\,px at 480-px width, threshold set post hoc) & excluded \\
\addlinespace[2pt]
Drift within a clip & held fraction changes & $36.30\%$ to $35.91\%$ by decile (Wilcoxon $p = 0.070$) & no detectable drift \\
\addlinespace[2pt]
Thermal build-up & rises with duration, recovers & none over 15.4--301.9\,s; $35.8\%$ after a 46-hour gap & not supported \\
\addlinespace[2pt]
Low light (Auto FPS) & affected clips darker & mean luminance $139.6$ (affected) vs $132.3$ (uniform) on a 0--255 scale (Mann--Whitney $p = 0.108$) & not supported \\
\addlinespace[2pt]
Visible to a builder & headers or frames differ & identical headers, metadata, codec; sharpness median $\approx 2500$ in both & not detectable \\
\addlinespace[2pt]
Tied to scene motion & hold density tracks flow & Spearman $+0.00$ (IQR $-0.07$ to $+0.13$, 11 clips) & no \\
\bottomrule
\end{tabular}
\end{center}
\end{table}

\subsection{Placement of held intervals}
\label{sec:placement}

\begin{figure}[!htb]
\begin{center}\includegraphics[width=0.80\linewidth]{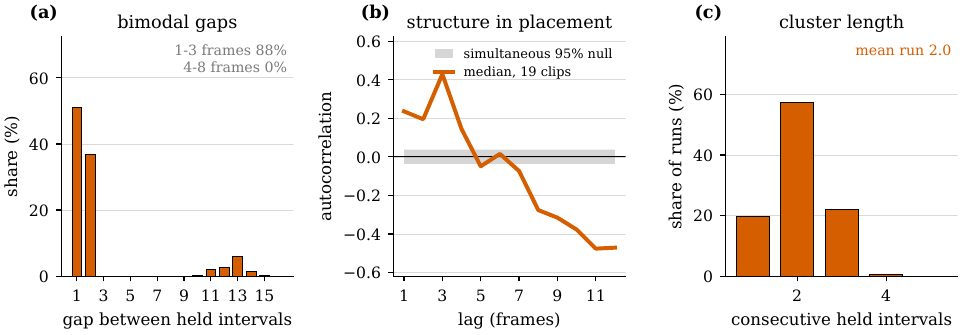}\end{center}
\caption{How held intervals are placed. (a)~Gaps between consecutive held
intervals, pooled over the affected population. Mass at one and two frames, a
second mode beyond ten, and nothing between: the signature of clustered
placement. (b)~Autocorrelation of the held-interval indicator: median across all
19 affected clips, against the simultaneous $95\%$ band of a permutation null
pooled over clips (grey), the band the significance tests use. (c)~Lengths of runs
of consecutive held intervals.}
\label{fig:placement}
\end{figure}

The holds are not scattered: they cluster, so whole averaging windows rather than
single frames are affected. Against a null that permutes each clip's held-interval
indicator 2{,}000 times, with a simultaneous band controlling the family-wise error
rate, all 19 clips show significant autocorrelation ($p = 0.0005$, the resolution
limit), positive at lags 1 to 3 (mean $+0.237$, $+0.196$ and $+0.424$, each surviving
Benjamini--Hochberg correction). Gaps between holds are bimodal
(Figure~\ref{fig:placement}a): $50.9\%$ are one frame, $36.7\%$ two, none four to
eight and $12.3\%$ ten or more, over-dispersed against a memoryless process
(standard deviation $3.70$ against $2.23$). Holds thus arrive in short clusters
separated by clean runs of ten to fourteen frames, which drives the autocorrelation
to $-0.47$ at lags 11 and 12 (Figure~\ref{fig:placement}b). The clusters recur with
a period of 22.6 frames (22.3 to 22.9 in every clip), about 128\,ms or 7.8\,Hz,
matching no documented device periodicity.

The rate does not drift (detrending reduces the timestamp residual by $0.0\%$), and
a constant target rate predicts the hold sequence no better than chance ($53.92\%$
against $54.07\%$). A 22.6-frame cadence carrying about eight holds reproduces both
the $35.8\%$ held fraction and the 176\,fps rate, so the behaviour is rate-driven per
cluster and irregular within one.

\subsection{Predictions against measurement}
\label{sec:predictions}

\begin{figure}[!htb]
\begin{center}\includegraphics[width=0.80\linewidth]{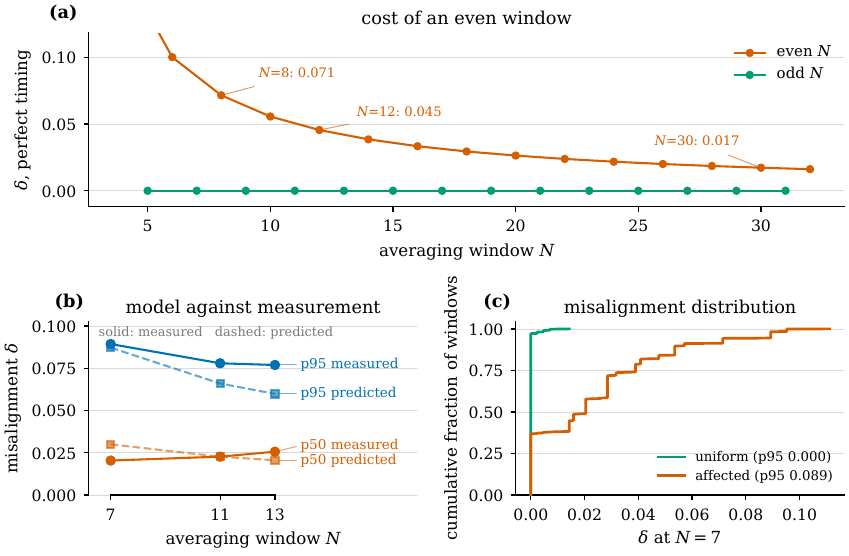}\end{center}
\caption{The model against the audit. (a)~Misalignment under perfect timing
(Eq.~\ref{eq:even}): zero for odd $N$, $1/(2(N-1))$ for even $N$. (b)~Measured
median and p95 misalignment in the affected population (solid) against P3 at
$p = 0.359$ (dashed), treating the timing term as Gaussian (median $0.674\sigma$,
p95 $1.96\sigma$): the scale agrees, but the median rises with $N$ where P3 predicts a fall, because
holds are not independent. (c)~Cumulative distribution of $\delta$ at $N = 7$: a point mass at
zero for uniform clips; up to $\delta \approx 0.1$ for affected ones.}
\label{fig:model}
\end{figure}

Figure~\ref{fig:model} sets the audit against Section~\ref{sec:model}. P1 holds to
the precision the clock allows: in the uniform population the misalignment is
$0.0000$ at the median for $N = 7$, 11 and 13, with a standard deviation of $0.0006$
to $0.0009$ from $\pm1$ tick dither, 25 to 37 times below the affected spread. P3 (Eq.~\ref{eq:energy}) recovers the order of magnitude from $p$ alone, at worst $22\%$ short at the 95th
percentile for $N=13$. Direct simulation reproduces the closed form to within $3.6$
to $6.4\%$, so the gap reflects its assumptions, and the median's trend in $N$ is reversed
because the independence it assumes fails. Measured over 118{,}848, 71{,}298 and 59{,}409 averaging windows of the affected
clips, $\sigma_\delta$ is $0.0255$, $0.0228$ and $0.0220$ at $N = 7$, 11 and 13. By
P4, at an illustrative blur length of 30 to 50 pixels this would be a label blur of $0.66$ to $1.28$ pixels, longer than our own affected
clips produce (Section~\ref{sec:scope}); an extrapolation, since the offset scales with
blur length. What this does to the labels built from these clips is the question of
Section~\ref{sec:consequences}.

\section{Consequences for supervision}
\label{sec:consequences}

The same violation has opposite consequences for the two kinds of label built from
high-speed video: blur labels averaged over held intervals cannot be repaired,
while interpolation labels can.

\subsection{Deblurring labels on this corpus}
\label{sec:scope}

The affected clips are the low-motion ones. Median dense flow per 240\,fps interval, pooled over all sampled intervals rather
than motion-selected windows, is $1.18$ pixels in the uniform population against
$0.14$ in the affected one (p95 $4.95$ against $1.91$). The consequence across this footage is small; since the offset scales with blur
length, benchmarks that select high-motion windows would see larger offsets, which
this corpus cannot measure. Transplanted onto the six longest uniform clips, the measured hold patterns put the blur label a median $0.19$ pixels (p95
$1.53$) from the true centroid at blur lengths of 10 to 12 pixels, against $0.75$
pixels for an even window on the same footage. The even-$N$ violation, by contrast, applies at any motion.

\subsection{Interpolation labels are repairable}
\label{sec:vfi}

For interpolation triplets at stride $k$ the middle frame is designated $t = 0.5$,
while its true normalised time is $t_{\text{true}} = (T_{i+k} - T_i)/(T_{i+2k} - T_i)$.
In the uniform population the two agree to tick dither (standard deviation $0.0012$
or less). In the affected population (Table~\ref{tab:vfi}), at stride 8, whose input frames are 16 nominal intervals apart (a 15\,fps pair), the
designated midpoint sits near normalised time $0.61$ at the 75th percentile. The
signed error still averages to zero, so aggregate reporting conceals it exactly as
in P2.

\begin{table}[!htb]
\caption{Interpolation label error $t_{\text{true}} - 0.5$ in the affected population
(uniform: median zero, p95 at most $0.0031$). The signed mean vanishes; the p95
coincides at strides 4 and 8, which reach the same attainable ratio, $t = 7/11$.}
\label{tab:vfi}
\begin{center}
\begin{tabular}{llrrrrr}
\toprule
Population & Stride & signed mean & sd & $|{\cdot}|$ p50 & p75 & p95 \\
\midrule
affected & 4  & $-2.0\times10^{-6}$ & 0.0680 & 0.0385 & 0.0833 & 0.1364 \\
affected & 8  & $-1.2\times10^{-5}$ & 0.0856 & 0.0833 & 0.1087 & 0.1364 \\
affected & 16 & $-1.0\times10^{-5}$ & 0.0346 & 0.0322 & 0.0455 & 0.0581 \\
\bottomrule
\end{tabular}
\end{center}
\end{table}

Unlike the blur label, the interpolation label is repairable: the true time
survives in the container, and a time-conditioned model can be given
$t_{\text{true}}$ instead of $0.5$ \citep{zhong2024clearer}. On real footage
carrying the measured hold patterns, doing so gains $1.41$\,dB at stride 8 for a
flow-based interpolator with the time-conditioned warping of
\citet{jiang2018superslomo}, better in $85\%$ of 277 triplets.

The blur label cannot be repaired this way: it is produced by averaging unevenly
spaced samples with equal weights, and no relabelling recovers an exposure that
never physically occurred. The remedy we can state is to synthesise only from
windows whose durations pass the timing check. Because held intervals cluster,
filtering keeps $29\%$ of 7-frame windows in affected footage, $11\%$ at $N = 11$
and $3\%$ at $N = 13$, and the surviving windows lie inside clean runs by
construction, so they are not a random sample of the footage.

\subsection{Screening a corpus before synthesis}
\label{sec:artifacts}

Both conditions can be checked before a single frame is averaged. Our screening
tool reads a QuickTime or MP4 sample-duration table in milliseconds without decoding. Durations pass when the linear-fit residual is at most $0.75$ ticks
($7.5\%$ of the nominal interval, so the test transfers to other timescales) and no
duration exceeds $1.5$ times the median, and parity is checked against the intended
$N$.

iPhone-HighFPS is a diagnostic instrument, not a training corpus: its
verified-uniform part is 26.8 minutes, far too small to compete with established
benchmarks \citep{choi2020cain,sim2021xvfi}. Subsampling a uniform clip by stride
$k$ gives the same scene at $240/k$\,fps with every container sample time known, and
displacement is linear in stride to within $4\%$ over a factor of 32. Scenes are
diverse (median nearest-neighbour cosine similarity between $32 \times 32$
thumbnails $0.554$, p90 $0.673$) and consistently sharp within each clip (coefficient of variation $0.02$ to $0.07$). The affected clips, with a container timestamp per frame, test interpolation at
arbitrary times.

The screening tool and analysis code are available at
\url{https://github.com/C-loud-Nine/Highfps_GT_Synthesis}. We will release iPhone-HighFPS upon publication: the 51 clips, partitioned by
population, with per-clip timing metadata. Redaction re-encodes the files but preserves and verifies each sample-duration table
and frame count; frame sizes and independent-clock values, which the copies cannot
reproduce, are given per clip.

\section{Limitations}
\label{sec:limitations}

The model assumes locally linear motion within the window, as the convention
itself does. We characterise the population-optimal predictor analytically and
test it only with a linear filter on synthetic scenes. A deep network may read direction cues from the blur, which would lower the hold-pattern cost and raise the parity cost as the double image
sharpens toward a shift. The interpolation gain likewise comes from one flow-based interpolator.

Measurements come from one device, one operating system version and two sessions,
with between-session spread in held fraction 53 times the mean within-session
spread, so the effective sample for session-level claims is two, not 51; the
released tool exists to test whether the findings transfer. We have not identified the mechanism: we find no support for thermal accumulation
or low light and exclude drift and fixed-rate decimation, but the trigger is not
observable from the files.
We do not quantify the effect on benchmarks built from this footage, including
iPhoneBlur \citep{shafi2026iphoneblur}, nor trace it
through other datasets, whose camera originals are rarely distributed.

Three remedies go untested: duration weighting cannot supply samples the sensor
never produced, a reference at $\bar{T}$ \citep{su2017dvd} reduces the offset
without evening the sampling, and misaligned-pair losses \citep{zhang2019zoom}
target spatial misregistration.
Extracted image pairs lose the sample-duration table, so the condition becomes
uncheckable after release, and rolling-shutter readout time is absent from this
metadata, so intra-frame timing is out of scope.

\section{Conclusion}

Frame averaging is sound for every capture timing only when the window is odd and the sample
durations within it are equal, and both can be checked before synthesis. None of the synthesis procedures we cite checks durations, and where parity is chosen deliberately it
is for other reasons. Three practices follow. Builders should choose an odd window and verify durations
from the container before averaging. Datasets should report the spread of label
offsets rather than their mean, since a zero mean certifies nothing. Extracted pairs
should carry their true timestamps, which repair interpolation labels and are
otherwise lost. Ground truth is only as valid as the timing it assumes.

\subsubsection*{Broader Impact Statement}

The corpus is public-street video in which passers-by and vehicle plates may
appear incidentally; no footage targets individuals. Released clips carry the video track only, with faces and plates blurred, location
metadata stripped and a removal route, under a permissive research licence.
The analysis describes a commercial product's behaviour as measured, without
attributing intent or defect.

\bibliography{main}
\bibliographystyle{tmlr}

\end{document}